\documentclass[10pt,twocolumn,letterpaper]{article}

\usepackage{cvpr}              

\usepackage[dvipsnames]{xcolor}

\definecolor{cvprblue}{rgb}{0.21,0.49,0.74}
\usepackage[pagebackref,breaklinks,colorlinks,citecolor=cvprblue]{hyperref}

\usepackage{url}

\usepackage{epsfig}
\usepackage{graphicx}
\usepackage{amsmath}
\usepackage{amssymb}

\def\paperID{1385} 
\def\confName{CVPR}
\def\confYear{2024}

\usepackage{pgf-pie}
\usepackage{subfloat}
\usepackage{multirow}
\usepackage{booktabs}
\usepackage{makecell}
\usepackage{mdframed}
\usepackage[accsupp]{axessibility}

\def\Dataset{{\textit C}-VQA}
\def\Real{{\textit C}-VQA-Real}
\def\Syn{{\textit C}-VQA-Synthetic}

\definecolor{darkgreen}{HTML}{228B22}
\definecolor{darkgray}{HTML}{757575}

\newcommand{\tablestyle}[2]{\setlength{\tabcolsep}{#1}\renewcommand{\arraystretch}{#2}\centering\small}
\newcommand{\tablestylesmaller}[2]{\setlength{\tabcolsep}{#1}\renewcommand{\arraystretch}{#2}\centering\footnotesize}
\newcolumntype{P}[1]{>{\centering\arraybackslash}p{#1}}

\usepackage[normalem]{ulem}

\newcommand{\authorskip}{\hspace{2.5mm}}
\newcommand{\affilationskip}{\hspace{5mm}}
\usepackage{tocloft}

\makeatletter
\renewcommand{\numberline}[1]{%
  \@cftbsnum #1\@cftasnum\hspace*{1em}\@cftasnumb%
}
\makeatother

\usepackage{listings}

\title{What If the TV Was Off? Examining Counterfactual Reasoning Abilities of Multi-modal Language Models}

\author{
\hspace{-2mm}Letian Zhang{\normalsize\textsuperscript{$1$}} \authorskip
Xiaotong Zhai{\normalsize\textsuperscript{$2$}} \authorskip
Zhongkai Zhao{\normalsize\textsuperscript{$3,6$}} \authorskip
Yongshuo Zong{\normalsize\textsuperscript{$4$}} \authorskip
Xin Wen{\normalsize\textsuperscript{$5,6$}} \authorskip
Bingchen Zhao{\normalsize\textsuperscript{$4,6$}} \vspace{2mm} \\
\normalsize
\textsuperscript{$1$}Tongji University \affilationskip
\textsuperscript{$2$}University of Warwick \affilationskip
\textsuperscript{$3$}National University of Singapore \\ \normalsize
\textsuperscript{$4$}University of Edinburgh \affilationskip
\textsuperscript{$5$}The University of Hong Kong \affilationskip
\textsuperscript{$6$}LunarAI
}

\begin{document}
\maketitle

\begin{abstract}

Counterfactual reasoning, a fundamental aspect of human cognition, involves contemplating alternatives to established facts or past events, significantly enhancing our abilities in planning and decision-making. 
In light of the advancements in current multi-modal large language models, we explore their effectiveness in counterfactual reasoning. 
To facilitate this investigation, we introduce a novel dataset, \Dataset, specifically designed to examine the counterfactual reasoning capabilities of modern multi-modal large language models. 
This dataset is constructed by infusing original questions with counterfactual presuppositions, spanning various types such as numerical and boolean queries. 
It encompasses a mix of real and synthetic data, representing a wide range of difficulty levels. 
Our thorough evaluations of contemporary vision-language models using this dataset have revealed substantial performance drops, with some models showing up to a 40\% decrease, highlighting a significant gap between current models and human-like vision reasoning capabilities. 
We hope our dataset will serve as a vital benchmark for evaluating the counterfactual reasoning capabilities of models.
Code and dataset are publicly available at \url{https://bzhao.me/C-VQA/}.
\end{abstract}

\section{Introduction}

\begin{figure}[t]
    \centering
    \includegraphics[width=1\linewidth]{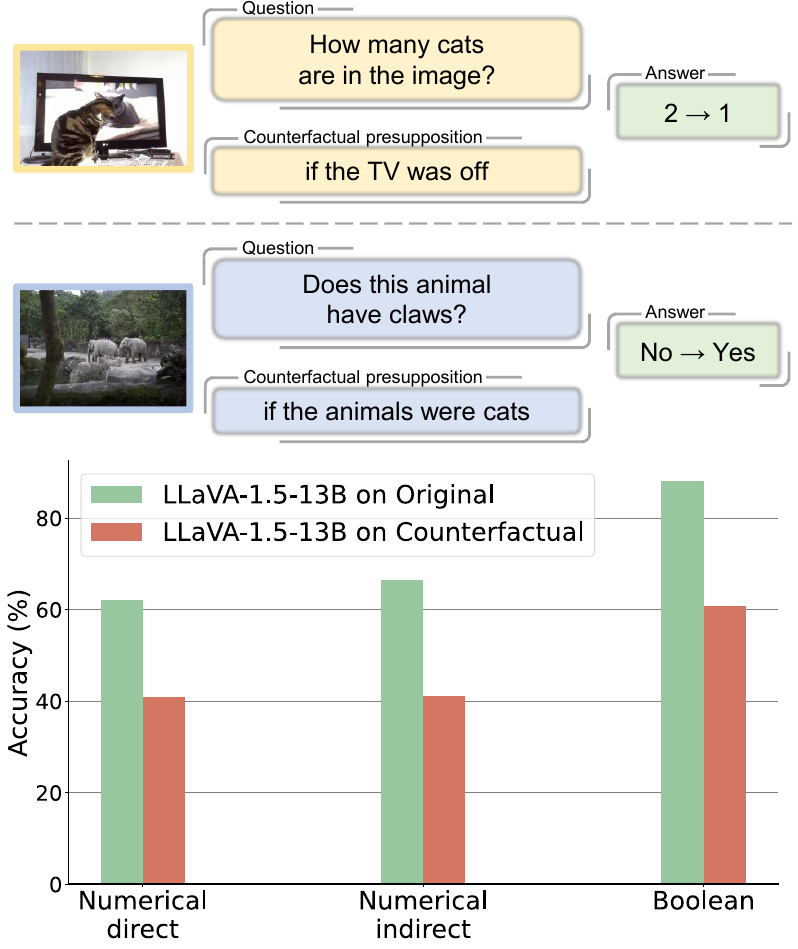}
    \vspace{-2em}
    \caption{
    \textbf{Examples of \Dataset (top), and performance comparison of LLaVA-1.5~\cite{liu2023improved} w/ and w/o counterfactuality (bottom)}. \Dataset~ is constructed by adding counterfactual presuppositions to the questions. We observe that state-of-the-art models all exhibited significant performance drops on the counterfactual questions.
    }
    \label{fig:overview}
    \vspace{-2em}
\end{figure}

\begin{quote}
``\textit{Counterfactuals are the building blocks of moral behavior as well as scientific thought.}'' \\
\raggedleft
--- Judea Pearl, \textit{The Book of Why}
\end{quote}

\noindent Counterfactual ability is a pivotal cognitive function in humans, enabling us to envision alternate realities and outcomes based on different choices or events. This capacity underpins our decision-making, moral reasoning, and problem-solving skills. Recent development of multi-modal large language models (MLLMs)~\cite{liu2023llava,yin2023survey,zhao2023vision,zong2023self} have dramatically improved the capabilities of image recognition, image-based dialogue, and language grounding, \etc~\cite{Gupta2022VisProg,surismenon2023vipergpt,koh2023fromage,li2023blip2,liu2023llava,ge2023planting,ge2023making,tu2023sight, Qwen-VL}. These developments raise the possibility of achieving higher levels of artificial intelligence. Consequently, we pose a critical question: \textit{are contemporary MLLMs equipped for counterfactual reasoning?}

Visual question answering is one of the central tasks to evaluate the ability of vision-language models. Current benchmarks focus on evaluating different aspects of the abilities, such as visual recognition (\eg, VQAv2~\citep{goyal2017making}), external knowledge (\eg, OKVQA~\citep{marino2019okvqa}), compositional reasoning (\eg, CLEVR~\citep{johnson2017clevr}). They only require the model to understand the image contents by grounding the concept involved in the language questions to the image content. With the help of large language models, these tasks are relatively easy to solve, as evidenced by the high scores achieved by modern MLLMs~\citep{liu2023llava, dai2023instructblip, chen2023minigptv2}. The most relevant benchmarks are CF-VQA~\citep{niu2021counterfactual} and VQA-CP~\citep{agrawal2018don}. However, they focus on constructing the counterfactual questions between the training set and the test set, \eg, showing a green banana in the test set while most of the bananas in the training set are yellow. They do not evaluate the counterfactual ability of the MLLMs - \textit{the skill to imagine ``what if" scenarios that differ from what actually happened}.

In light of this, we propose a novel and challenging evaluation scenario where the language model not only needs to query for the correct visual representation of image content but also needs to be able to perform counterfactual reasoning on those representations.
This cannot be solved by encoding world knowledge or object spatial relation alone, but also requires the model to further understand and imagine the given scenarios.
Specifically, in our proposed benchmark, each visual question is modified by a counterfactual presupposition - making an assumption based on a scenario that did not actually happen, but could have. This requires the MLLMs to understand the visual content and then perform reasoning over the image contents accordingly to answer the questions.
For example, in~\cref{fig:overview}, to answer the question of ``How many cats are in the image?", the model can directly query the image and count the occurrence of the visual concept ``cat''. 
However, with the counterfactual presupposition ``if the TV was off'', one of the cats on the screen will not be visible. Thus, it poses a more challenging scenario for the models as the model not only needs to identify the occurrence of the concept ``cat'', but also reason about its state after a counterfactual operation is taken.

We build our dataset on top of the commonly used VQAv2 dataset~\cite{antol2015vqa}, where we collect over $3,000$ images suitable for asking counterfactual questions. 
In addition, we also created over $3,000$ synthetic images and corresponding questions to further test the out-of-distribution ability of multi-modal large language models.


We evaluate a wide array of comtemperory state-of-the-art multi-modal language models on our dataset, such as LLaVA~\cite{liu2023llava}, MiniGPT4~\cite{zhu2023minigpt,chen2023minigptv2}, BLIP2~\cite{li2023blip2},  InstructBLIP~\cite{dai2023instructblip}, Qwen-VL~\cite{Qwen-VL} and CogVLM~\cite{wang2023cogvlm}.
Our experiments show several interesting findings: (1) Neuro-symbolic models perform worse than end-to-end models on complex counterfactual reasoning; (2) No model family can consistently address counterfactual questions. All of the models suffer from a large performance drop from our counterfactual questions; (3) Even the strongest GPT-4V model~\cite{openai2023gpt4v} cannot solve our benchmark; (4) The MLLMs also demonstrate a systematic bias in answering gender-related counterfactual questions.

We summarize the main contributions of this paper:
\begin{enumerate}
    \item We propose a novel and challenging dataset~\Dataset~with both real and synthetic image-question pairs. The questions contain counterfactual presuppositions and are of various difficulties.
    \item We conduct extensive evaluation of current state-of-the-art vision-language models on our dataset. We show that they struggle with counterfactual reasoning.
    \item We summarize our findings from the experiments on proposed~\Dataset, hoping to provide valuable insights for future research in this field.
\end{enumerate}


\begin{figure}[t]
    \centering
    \includegraphics[width=1\linewidth]{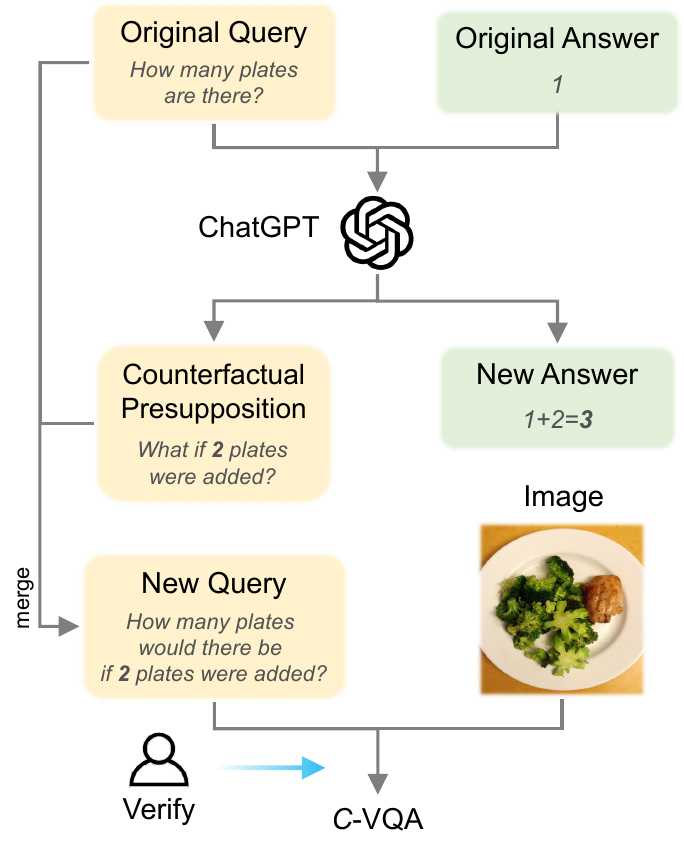}
    \vspace{-2.5em}
    \caption{\textbf{Our annotation flow for~\Real}. We select images and questions from the VQAv2 dataset~\cite{antol2015vqa}, and then utilize ChatGPT to add counterfactual presupposition to the questions and get the corresponding answers. All questions and answers are carefully inspected by human annotators.}
    \label{fig:annotation}
\end{figure}

\section{Related Works}

\paragraph{Visual Question Answering.}
Visual question answering (VQA) aims to evaluate machines' capabilities of visual understanding, visual reasoning, and the application of commonsense knowledge.
Several datasets have been proposed for VQA~\cite{antol2015vqa,goyal2017making,hudson2019gqa,marino2019okvqa,ren2015exploring,johnson2017clevr,cherian2023deep,li2022from,wu-etal-2023-acquired}.
COCO-QA~\cite{ren2015exploring} and VQA~\cite{antol2015vqa} are the first to propose the task of VQA, and they contain enormous pictures and questions covering daily life objects. 
Builds on the VQA dataset. VQAv2~\cite{goyal2017making} presents a more rigorous evaluation for the VQA task by mitigating language biases and shortcuts in the VQA dataset.
Further works~\cite{marino2019okvqa,johnson2017clevr,hudson2019gqa} continue to extend VQA evaluation to different aspects of image understanding.
CLEVR dataset~\cite{johnson2017clevr} devises a pipeline for generating synthetic data for evaluating the compositional reasoning ability of VQA models.
GQA dataset~\cite{hudson2019gqa} leverages the scene graph structure to generate reasoning questions on real-world images to test the compositional reasoning ability.
In this work, we build our dataset~\Dataset~on the top of VQAv2~\cite{goyal2017making}, where the questions and answers are modified by counterfactual presupposition. Different from previous VQA tasks, our proposed counterfactual dataset provides a new challenging scenario for complementary MLLMs. It further examines the models' abilities to parse the scene structure and reason about the observed world after the counterfactual presuppositions.

\vspace{-.5em}
\paragraph{Evaluation of Reasoning Abilities.}
The evaluation of generative models has always been a difficult and challenging problem, as well as for large language models.
There have been many efforts for the evaluation of the reasoning ability of LLMs in different aspects~\cite{GSM8k, Hendrycks2021MMLU, Hendrycks2021MATH, Yu2022ifqa, wu2023reasoning, niu2021counterfactual, agrawal2018don}. For example, GSM8k~\cite{GSM8k} and MATH~\cite{Hendrycks2021MATH} evaluate the mathematical reasoning of LLMs, while MMLU~\cite{Hendrycks2021MMLU} aggregates a diverse range of subjects and tasks for evaluation.
For counterfactual reasoning, IfQA~\cite{Yu2022ifqa} proposes the first QA dataset that specifically designed to assess the counterfactual reasoning capabilities of language models.
In~\cite{wu2023reasoning}, the counterfactual reasoning abilities of state-of-the-art LLMs~(GPT-4~\cite{openai2023gpt4}, Claude~\cite{claude}) are evaluated under eleven different tasks with counterfactual presuppositions. It is shown that current LLMs cannot reason with counterfactuals reliably.
Most related to our work is CF-VQA~\cite{niu2021counterfactual} and VQA-CP~\cite{agrawal2018don} which focus on testing the ability of VQA models to answer ``counterfactual'' questions which are defined as having different property distribution in training and testing set, such as training with yellow bananas but test with green ones.
In our work, we propose to build the first visual question-answering dataset with counterfactual presuppositions directly added to the question text, this enables the testing of current SOTA MLLMs which leverages internet-scale training data. 
Our evaluation results show current approaches for MLLMs do not facilitate counterfactual reasoning. Thus, further work should be done to create stronger MLLMs.

\vspace{-.5em}
\paragraph{Multi-Modal LLM Benchmarks}
With the increasing interest in training novel multi-modal LLMs that can perceive multi-modal inputs, various benchmarks for evaluating the performance of MLLMs are proposed. 
MME~\cite{fu2023mme} measures the performance of multi-modal LLMs using both perception and cognition abilities on $14$ subtasks.
MM-Bench~\cite{liu2023mmbench} covers over $20$ different ability dimensions. A robust evaluation method is also proposed that leverages ChatGPT to match the model prediction to given choices.
SEED-Bench~\cite{li2023seedbench} proposes a benchmark with over $19$k questions, covering both image and video modality to evaluate the performance of multi-modal LLMs.
MM-Vet~\cite{yu2023mmvet} defines $16$ emergent tasks of interest from $6$ core vision-language capabilities, with an LLM-based evaluator, relative strengths and weaknesses of different system paradigms are identified.
In this paper, we propose an orthogonal direction for evaluating the reasoning ability of current multi-modal LLMs: counterfactual reasoning.

\begin{table*}[t]
\caption{
\textbf{Three instances in \Syn.} They are made from different templates. 
}
\vspace{-1em}
\centering
\begin{tabular}{p{5.2cm} p{5.2cm} p{5.2cm}}
 
 \begin{minipage}[b]{0.6\columnwidth}
    \centering
    \raisebox{-.5\height}{\includegraphics[width=\linewidth]{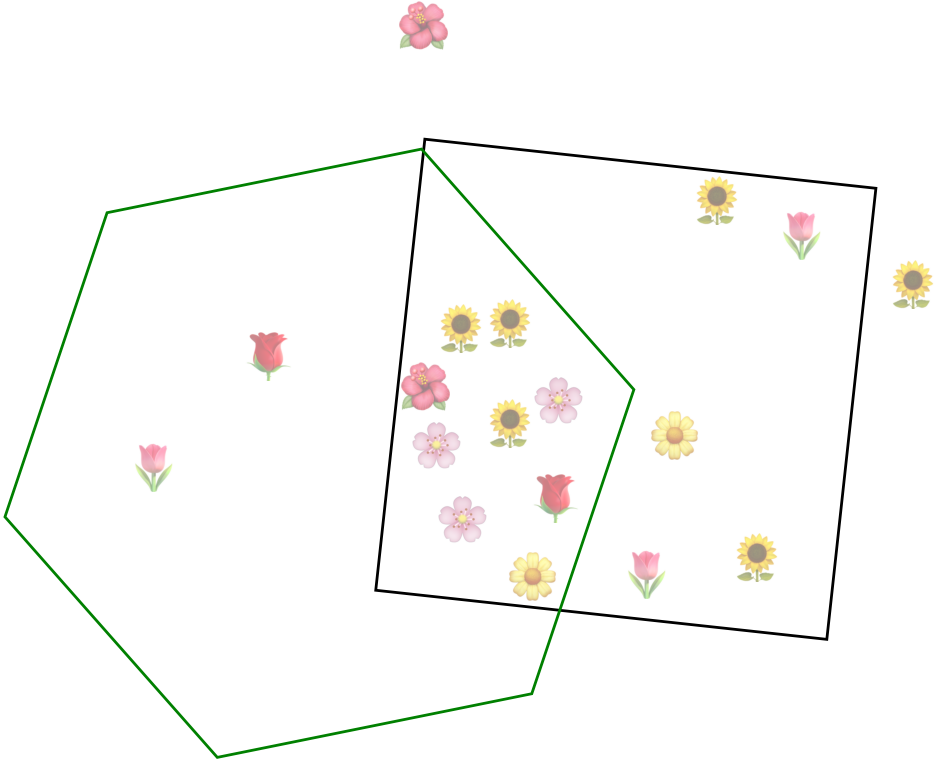}
    }
\end{minipage}
&
\begin{minipage}[b]{0.6\columnwidth}
    \centering
    \raisebox{-.5\height}{\includegraphics[width=\linewidth]{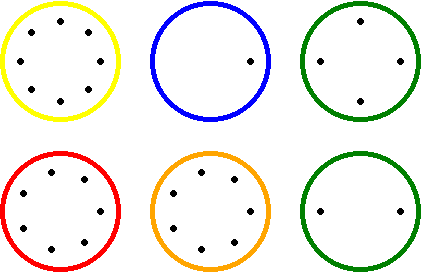}}
\end{minipage} 
& 
\begin{minipage}[b]{0.6\columnwidth}
    \centering
    \raisebox{-.5\height}{\includegraphics[width=\linewidth]{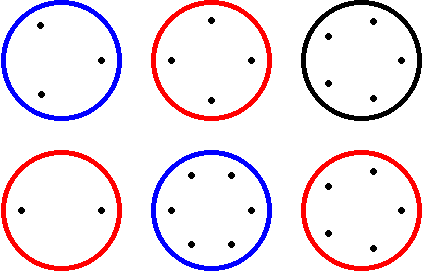}}
\end{minipage} 
\\
\small{How many flowers would be inside black polygons if all flowers in green polygons were removed?
Select the correct answer: A:5  B:1  C:3  D:7  }
& 
\small{How many dots would there be in all the circles together if 24 dots were removed from the circles?
Select the correct answer: A:7  B:11  C:3  D:9}
& 
\small{How many dots would a circle contain at most if one of the circles with most dots were removed?
Select the correct answer: A:5  B:1  C:3  D:7 }
\\
\end{tabular}

\label{tab:abstract}
\end{table*}

\section{Dataset}
This section presents the construction process of our proposed dataset~\Dataset, which consists of two parts: \Real~and~\Syn.
\Real~contains $3,144$ image and question-answer pairs, where each question is not only related to the image content but also comes with a counterfactual presupposition.
All images and original questions of \Real~come from the VQAv2 dataset.
\Syn~contains $3,000$ image and question-answer pairs, and all images and questions in \Syn~are generated automatically. 
Each image corresponds to an original question and a counterfactual question.
The counterfactual presuppositions in \Dataset~enable a new and challenging scenario for VQA models.
Both \Real~and \Syn~contain numerical and boolean questions that can be answered with a number or a boolean value respectively.
We illustrate the counterfactual presupposition generation and verification steps below.

\subsection{Annotation}

\paragraph{Counterfactual presupposition type of \Real.}
When designing counterfactual questions, we employ different types of counterfactual presuppositions as follows.

\noindent\textbf{Numerical questions} are split into two groups: direct group and indirect group.
In the \textbf{numerical direct} group, we add counterfactual presuppositions that change the original answers directly. 
For example, these questions typically have the form \textit{``How many X would there be if two X were added/removed?"}.
In the \textbf{numerical indirect} group, the counterfactual presuppositions change the original answers indirectly. 
It requires more reasoning steps to get the new answers.
For example, answering the sentence \textit{``How many cats would be there if the TV was off?"} requires the model to recognize how many cats are images on the TV and understand that when the TV is off, those cats will no longer be there.
For \textbf{boolean questions}, the counterfactual presuppositions are often designed to reverse the fact as well as the answer. 
For example, \textit{``Would the cat be asleep if it was woken up?"}.
Examples of these counterfactual modifications are presented in~\cref{fig:overview}.

\vspace{-.5em}
\paragraph{Question and answer annotation of \Real.}
A two-stage annotate-and-prompt process is employed to create \Dataset.
First, we manually annotate $200$ questions and answers for each of groups we have defined above. 
Then, these manually annotated questions and answers are used as in-context-examples~\cite{dong2023survey} to prompt ChatGPT~\cite{openai2023gpt4} to generate new counterfactual modified questions for the remaining samples.
Additionally, to maximize the correctness of ChatGPT-generated questions, we leverage chain-of-thought~\cite{wei2023chainofthought} and divide the whole task into several smaller tasks: read the original question, figure out a proper counterfactual presupposition, figure out how the answer will change, and output the new question and answer. 
We also provide in-context examples to further ensure the sentence structure and to illustrate various counterfactual presuppositions.
For numerical question groups, we first prompt ChatGPT to produce a counterfactual presupposition and then generate a new question with it, and the new answer is also calculated.
The annotation flow is plotted in~\cref{fig:annotation}. 
However, this strategy cannot be directly applied to boolean questions. 
We notice the randomly generated counterfactual presuppositions often fail to flip the original answer, and thus a large proportion of answers remain unchanged or even cannot be determined.
Therefore, we employ a prompting strategy that flips the original answer first and then generates a corresponding counterfactual presupposition.

\subsection{Verification}


As we leverage ChatGPT in the process of generating the annotation of \Real, we employ a rigorous verification process to ensure the correctness of the generated annotations in order to remove any hallucination or calculation mistakes\footnote{At the time of curating the dataset, GPT-4V API was not publicly available yet~\cite{openai2023gpt4v}.}.  
Therefore, we further verify the questions and answers manually to address the errors caused by ChatGPT.
Therefore, although the counterfactual questions generated by ChatGPT are idiomatic, they may be improper in the context of the scene in the image.
Thus we further verify the questions and answers manually to address any potential errors.
Our verification mainly consists of two stages: (i) image-related verification and (ii) answer-reasonability verification.
And each image is at least verified by two people to ensure correctness. We examine the following two key points:
\paragraph{(i) Whether the new question is image-related?}
To make sure that the generated question is indeed asking the model to reason with the scene in the image, we manually examine all generated counterfactual questions and remove all questions that were modified to be not related to the image content.
\paragraph{(ii) Whether the new answer is reasonable?} 
Automatically generated answers may be wrong since ChatGPT may make calculation mistakes, reasoning mistakes, etc.
We correct these answers manually.
Furthermore, some questions may be ambiguous, with no deterministic answers. 
We remove these questions from the dataset.
Since ChatGPT is text-only and cannot see the images, it has to access to the position or color of the objects.
Thus, it tends to add counterfactual presuppositions that remove all objects rather than just manipulate a few of the objects.
To ensure a rigorous evaluation, we manually annotate all numerical indirect questions.
After completing all the annotations, we use ChatGPT again to inspect and polish all the questions again in order to make sure the questions are grammatically correct.


\subsection{Implementation of \textbf{\Syn}} 

In this section, we introduce the procedure we take to generate the \Syn~which contains synthetic images with counterfactual questions and answers.
The advantage of \Syn~is that this enables the automatic generation of high-quality counterfactual image-related questions, saving the time and labor cost of manually annotating.
We designed six abstract question types in total and generated $500$ instances for each type, resulting in a total of $3,000$ synthetic images with counterfactual questions. 
All images are generated via a predefined procedure in~\cite{cherian2023deep} randomly, and we use templates to produce original and counterfactual questions. 
Both original and counterfactual questions are in a multiple-choice question format. 
\cref{tab:abstract} shows three instances of \Syn.
In the following, we detail the design of our \Syn~by explaining the process of generating two sets of synthetic images and questions: Flower-counting and Dot-counting.

\begin{table}[t]
\caption{\textbf{Six question templates of \Syn.} Each template has 500 instances with different images and questions. \{\texttt{Color}\} and $N$ are randomly generated for each instance.}
\vspace{-.8em}
\centering
\tablestylesmaller{5pt}{1}
\begin{tabular}{P{.6cm} p{7cm}}
\toprule
Type  & Question Template  \\ 
\midrule
\multirow{7}{*}{\rotatebox[origin=c]{90}{Flower-counting}}
& How many flowers would be outside the \{\texttt{Color}\} polygons if all polygons were \{\texttt{Color}\}?  \\ 
\cmidrule(r){2-2} 

& How many flowers would be inside \{\texttt{Color}\} polygons if we removed $N$ flowers in \{\texttt{Color}\} polygons?  \\ 
\cmidrule(r){2-2} 

& How many flowers would be inside \{$\texttt{Color}_\texttt{1}$\} polygons if all flowers in \{$\texttt{Color}_\texttt{2}$\} polygons were removed?   \\ 
\midrule

\multirow{8}{*}{\rotatebox[origin=c]{90}{Dot-counting}}
& How many dots would there be in all the circles together if $N$ dots were removed from the circles?   \\
\cmidrule(r){2-2}

& How many dots would there be in the top three circles together if the two rightmost circles and dots in them were removed from the circles? \\ 
\cmidrule(r){2-2}

& How many dots would a circle contain at most if one of the circles with most dots were removed? \\ 
\bottomrule
\end{tabular}
\label{tab:template}
\vspace{-2em}
\end{table}

\vspace{-.5em}
\paragraph{Flower-Counting Puzzles.}
Three types of questions in \Syn~are based on flower-counting tasks. 
We first randomly sample two polygons $S_1$ and $S_2$ in different colors $C_1$ and $C_2$. 
The two polygons divide the whole image into four parts: $ \texttt{in}_1 \& \texttt{in}_2 $, $ \texttt{in}_1 \& \texttt{out}_2 $, $ \texttt{out}_1 \& \texttt{in}_2 $ and $ \texttt{out}_1 \& \texttt{out}_2 $, where $ \texttt{in}_i $ means inside both $ S_i $, and $ \texttt{out}_i $ means outside $ S_i $.
Then, we select several flower instances from the Icons-50 dataset~\cite{hendrycks2019benchmarking} and insert them randomly into the four parts. 
The flowers are guaranteed not to intersect with the edge of polygons to ensure an accurate answer for reasoning questions. 
The number of flowers in part $p$ is denoted as $\texttt{num}(p)$.
With the images procedurally generated, we design questions related to the image contents by filling in question templates.
The question templates are shown in~\cref{tab:template}.
To answer those questions well, a multi-modal model not only needs to reason about the spatial relationship, but also needs to perform counterfactual reasoning.
For example, the first question template would require the model to reason about the color and the spatial relationship of the image as well as the counterfactual change of color, and the third question template requires the model to reason about the intersection of regions and the counterfactual removal of items.

\begin{figure}[t]
	\centering
	\subfloat[Direct Group]{
		\begin{minipage}[b]{0.22\textwidth}
			\includegraphics[width=1\textwidth]{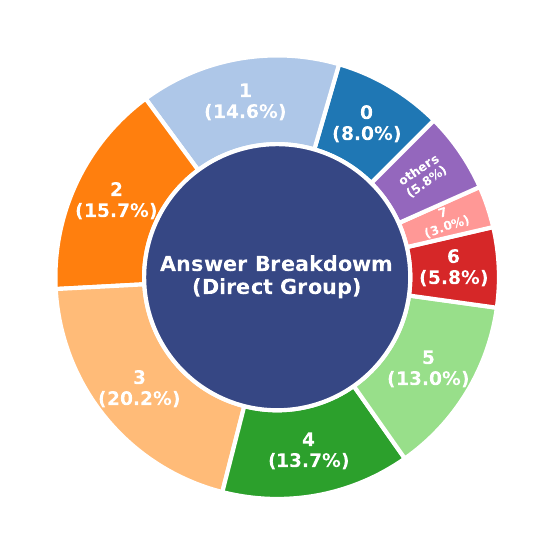}
		\end{minipage}
		\label{fig:direct}
    }
        \subfloat[Indirect Group]{
    		\begin{minipage}[b]{0.22\textwidth}
   		 	\includegraphics[width=1\textwidth]{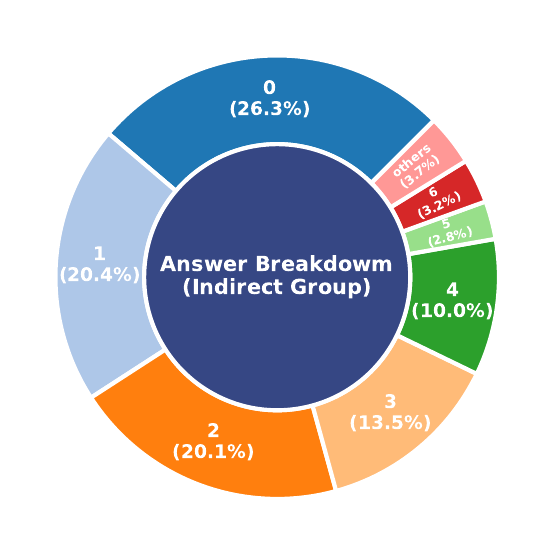}
    		\end{minipage}
		\label{fig:indirect}
        }
    \vspace{-.8em}
	\caption{
        \textbf{Breakdown of answers in numerical groups of \Real.}
        We show the percentage of answers in the numerical direct group and numerical indirect group.
        The share of 0, 1, and 2 in the indirect group are higher while the others are lower.
        }
	\label{fig:num_ans}
\end{figure}

\begin{table*}[t]
\caption{\textbf{Evaluation results on three subgroups of \Real~}by ViperGPT\cite{surismenon2023vipergpt}, VisProg\cite{Gupta2022VisProg}, BLIP2~\cite{li2023blip2},  InstructBLIP~\cite{dai2023instructblip}, MiniGPT4~\cite{zhu2023minigpt,chen2023minigptv2}, LLaVA~\cite{liu2023llava}, CogVLM~\cite{wang2023cogvlm}, and Qwen-VL~\cite{Qwen-VL}. All of them suffer from significant performance drop from counterfactual questions.
}
\vspace{-.8em}
\centering
\tablestyle{4.6pt}{1.05}
\begin{tabular}{llcccccc}
\toprule
\multirow{2}{*}{Model Type} & \multirow{2}{*}{Model} & \multicolumn{2}{c}{Numerical Direct} & \multicolumn{2}{c}{Numerical Indirect} & \multicolumn{2}{c}{Boolean} \\

\cmidrule(r){3-4} \cmidrule(r){5-6} \cmidrule(r){7-8}
& & Original & Counterfactual & Original  & Counterfactual & Original  & Counterfactual  \\
\midrule
\multirow{2}{*}{Neuro-symbolic}  
& VisProg  & 40.3 & 39.9$\,_\text{\bf\textcolor{BrickRed}{(-0.4)}}$ & 38.4 & 16.8$\,_\text{\bf\textcolor{BrickRed}{(-21.6)}}$ & 75.4 & 29.5$\,_\text{\bf\textcolor{BrickRed}{(-45.9)}}$ \\
& ViperGPT  &  83.8 &  71.4$\,_\text{\bf\textcolor{BrickRed}{(-12.4)}}$ & 78.6
 & 30.2$\,_\text{\bf\textcolor{BrickRed}{(-48.4)}}$ & 95.0  & 28.2$\,_\text{\bf\textcolor{BrickRed}{(-66.8)}}$ \\
\midrule
\multirow{12}{*}{End-to-end}

& BLIP2 (FlanT5$_{\rm XXL}$)  & 43.4 & 32.3$\,_\text{\bf\textcolor{BrickRed}{(-11.1)}}$ & 49.1 & 37.5$\,_\text{\bf\textcolor{BrickRed}{(-11.6)}}$ & 77.8 & \textbf{75.1}$\,_\text{\bf\textcolor{BrickRed}{(-2.7)}}$ \\ 

& InstructBLIP (FlanT5$_{\rm XXL}$) & 57.7  & \textbf{42.1}$\,_\text{\bf\textcolor{BrickRed}{(-15.6)}}$  & 58.9 & \textbf{43.8}$\,_\text{\bf\textcolor{BrickRed}{(-15.1)}}$ & 83.1  & 73.9$\,_\text{\bf\textcolor{BrickRed}{(-9.2)}}$ \\

& InstructBLIP (Vicuna-7B) & 64.5 & 30.1$\,_\text{\bf\textcolor{BrickRed}{(-34.4)}}$ & 64.2 & 30.2$\,_\text{\bf\textcolor{BrickRed}{(-34.0)}}$ & 83.9 & 50.4$\,_\text{\bf\textcolor{BrickRed}{(-33.5)}}$ \\  

& InstructBLIP (Vicuna-13B) & 63.5 & 41.7$\,_\text{\bf\textcolor{BrickRed}{(-21.8)}}$ & 63.4 & 39.7$\,_\text{\bf\textcolor{BrickRed}{(-23.7)}}$ & 86.6 & 61.8$\,_\text{\bf\textcolor{BrickRed}{(-24.8)}}$ \\
\cmidrule(r){2-8}

& MiniGPT-4  (Vicuna-7B) & 31.2 & 19.6$\,_\text{\bf\textcolor{BrickRed}{(-11.6)}}$ & 30.6 & 19.6$\,_\text{\bf\textcolor{BrickRed}{(-11.0)}}$ & 55.8 &  41.2$\,_\text{\bf\textcolor{BrickRed}{(-14.6)}}$ \\  

& MiniGPT-v2  (Llama2-Chat-7B) &55.0 & 25.9$\,_\text{\bf\textcolor{BrickRed}{(-29.1)}}$ & 55.2 & 29.1$\,_\text{\bf\textcolor{BrickRed}{(-26.1)}}$ & 76.8 &  46.7$\,_\text{\bf\textcolor{BrickRed}{(-30.1)}}$ \\  

\cmidrule(r){2-8}
& LLaVA-7B (Vicuna-7B) & 38.8 & 38.8$\,_\text{\bf\textcolor{BrickRed}{(-0.0)}}$  & 42.0 & 37.5$\,_\text{\bf\textcolor{BrickRed}{(-4.5)}}$ & 60.7 & 55.9$\,_\text{\bf\textcolor{BrickRed}{(-4.8)}}$ \\

& LLaVA-13B (Vicuna-13B) & 31.2 & 31.2$\,_\text{\bf\textcolor{BrickRed}{(-0.0)}}$  & 38.3 & 31.4$\,_\text{\bf\textcolor{BrickRed}{(-6.9)}}$ & 67.3 & 64.3$\,_\text{\bf\textcolor{BrickRed}{(-3.0)}}$ \\

& LLaVA-1.5-7B (Vicuna-7B) & 60.4 & 36.5$\,_\text{\bf\textcolor{BrickRed}{(-23.9)}}$  & 62.2 & 37.4$\,_\text{\bf\textcolor{BrickRed}{(-24.8)}}$ & 86.2 & 58.5$\,_\text{\bf\textcolor{BrickRed}{(-27.7)}}$ \\

& LLaVA-1.5-13B (Vicuna-13B) &62.0 &  41.0$\,_\text{\bf\textcolor{BrickRed}{(-21.0)}}$  & 66.4 & 41.2$\,_\text{\bf\textcolor{BrickRed}{(-25.2)}}$ &  88.0 & 60.7$\,_\text{\bf\textcolor{BrickRed}{(-27.3)}}$ \\

\cmidrule(r){2-8} 
& CogVLM-Chat & 49.2 &  20.3$\,_\text{\bf\textcolor{BrickRed}{(-28.9)}}$  & 50.3 & 25.9$\,_\text{\bf\textcolor{BrickRed}{(-24.4)}}$ & 83.6  & 60.0$\,_\text{\bf\textcolor{BrickRed}{(-23.6)}}$ \\

\cmidrule(r){2-8} 
& Qwen-VL-Chat & \textbf{65.1} &  30.7$\,_\text{\bf\textcolor{BrickRed}{(-34.4)}}$  & \textbf{69.2} & 29.4$\,_\text{\bf\textcolor{BrickRed}{(-39.8)}}$ & \textbf{88.1}  & 49.3$\,_\text{\bf\textcolor{BrickRed}{(-38.8)}}$ \\

\bottomrule
\end{tabular}

\label{tab:result}
\end{table*}

\begin{table}[ht]
\caption{\textbf{Evaluation results on~\Real~}by four end-to-end models when chain-of-thought prompt ``Let's think step by step:" is post-pended to the questions.} 
\label{tab:cot}
\vspace{-.8em}
\centering
\tablestylesmaller{1.7pt}{1.05}
\begin{tabular}{lcccccc}
\toprule
\multirow{2}{*}{Model} & \multicolumn{2}{c}{Num. Direct} & \multicolumn{2}{c}{Num. Indirect} & \multicolumn{2}{c}{Boolean} \\

\cmidrule(r){2-3} \cmidrule(r){4-5} \cmidrule(r){6-7}
& Ori. & Counterfact & Ori.  & Counterfact & Ori.  & Counterfact  \\

\midrule
InstructBLIP-13B & 63.2 & 34.2$\,_\text{\tiny\bf\textcolor{BrickRed}{(-29.0)}}$ & 62.3 & 33.7$\,_\text{\tiny\bf\textcolor{BrickRed}{(-28.6)}}$ & 82.8 & 51.0$\,_\text{\tiny\bf\textcolor{BrickRed}{(-31.8)}}$ \\

MiniGPT-v2-7B &46.8 & 21.4$\,_\text{\tiny\bf\textcolor{BrickRed}{(-25.4)}}$ & 44.0 & 24.5$\,_\text{\tiny\bf\textcolor{BrickRed}{(-19.5)}}$ & 72.9 &  46.3$\,_\text{\tiny\bf\textcolor{BrickRed}{(-26.6)}}$ \\ 

LLaVA-1.5-13B &62.1 &  41.7$\,_\text{\tiny\bf\textcolor{BrickRed}{(-20.4)}}$  & 64.9 & 42.2$\,_\text{\tiny\bf\textcolor{BrickRed}{(-22.7)}}$ &  87.9 & 60.9$\,_\text{\tiny\bf\textcolor{BrickRed}{(-27.0)}}$ \\

Qwen-VL-Chat & 65.2 &  29.7$\,_\text{\tiny\bf\textcolor{BrickRed}{(-35.5)}}$  & 68.4 & 24.9$\,_\text{\tiny\bf\textcolor{BrickRed}{(-43.5)}}$ & 87.4  & 47.5$\,_\text{\tiny\bf\textcolor{BrickRed}{(-39.9)}}$ \\

\bottomrule
\end{tabular}
\vspace{-2em}
\end{table}

\vspace{-.5em}
\paragraph{Dot-Counting Puzzles.}
The other three types of questions are based on dot-counting tasks. 
We arrange six circles in the images and then insert dots into these circles randomly. 
The questions are about the number of dots in the circles. 
For example, the first template is \textit{``How many dots are there in all the circles together?"} for the original question and \textit{``How many dots would there be in all the circles together if $N$ dots were removed from the circles?"} for the counterfactual question. 
Suppose the original answer is $ \texttt{ans}_1 $, we can calculate the counterfactual answer $ \texttt{ans}_2$ by the equation $ \texttt{ans}_2 = \texttt{ans}_1 - N $.
For the third type of counterfactual question, \textit{``How many dots would a circle contain at most if one of the circles with most dots were removed?"}, in order to fully evaluate the reasoning ability of models, the circle with the most dots is designed to be unique to ensure the counterfactual answer varies from the original one.

We present all six question templates in~\cref{tab:template}. All templates are ``How many" questions, and we insert random elements in images and templates. 
The answers are designed to have a wide range so that we can rigorously test the counting and reasoning ability of MLLMs. 
Additionally, the domain gap between these synthetic images poses a challenge for the current multi-modal models as \Syn~is asking for both domain generalization and counterfactual reasoning, which we argue is an important ability for multi-modal models.


\subsection{Dataset Statistics}
\paragraph{Question Type and Length.}
\Real~contains $3144$ questions in total, with $2014$ numerical questions and $1130$ boolean questions.  
In numerical questions, $1150$ questions are from the direct group and $864$ questions are from the indirect group.
All the numerical questions are \textit{``How many"} questions. 
Most of the questions in the boolean group start with ``Is" or ``Are" before they are changed into counterfactual ones. 
The average length of questions in \Real~is $13.15$ words, much longer than that of original questions ($5.88$ words)~\cite{antol2015vqa}. \Syn~contains $3000$ questions with a multiple-choice answer selection setting.
The average length of counterfactual questions in \Syn~is $25.33$ words, also longer than that of original questions ($16.00$ words).

\paragraph{Answer Statistics.}
For \Real, each answer in the numerical type is an exact number, with no ambiguous answers such as ``a lot" or ``many".
The distribution of the answers is shown in~\cref{fig:num_ans}. 
Each answer in the boolean type is a single ``yes" or a single ``no". 
The percentage of ``no" is $55.13\%$, while ``yes" is $44.87\%$. 
For \Syn, the options are randomly generated, and we control the proportion of each option close to $25\%$.


\section{Experiments}


In this section, we provide the evaluation of current state-of-the-art multi-modal LLMs on our proposed \Dataset.
Our evaluation covers both neuro-symbolic models like ViperGPT~\cite{surismenon2023vipergpt} and VisProg~\cite{Gupta2022VisProg} and end-to-end models including LLaVA~\cite{liu2023llava}, MiniGPT4~\cite{zhu2023minigpt,chen2023minigptv2}, BLIP2~\cite{li2023blip2},  InstructBLIP~\cite{dai2023instructblip}, CogVLM~\cite{wang2023cogvlm}, and Qwen-VL~\cite{Qwen-VL}.
Our evaluation results reveal several interesting findings as detailed below.
Implementation details are in the appendix.

\subsection{Experiments on \bf{\Real}}
\paragraph{Performance of End-to-end Models.}
As shown in~\cref{tab:result}, end-to-end models perform significantly worse with counterfactual questions than the original ones in all three groups. 
We find that reasoning difficulties have a significant impact on the results. 
For the numerical direct group, models require only one simple reasoning step to get the answer.
As a consequence, the difference between the evaluation accuracy of the original and counterfactual questions is smaller compared to other groups. 
We further evaluate four end-to-end models on~\Real~with a prompt ``Let's think step by step" for chain-of-thought reasoning~\cite{kojima2022large, wei2023chainofthought}. 
As shown in~\cref{tab:cot}, the performance of the models remains constant or even drops with CoT, which indicates that the CoT strategy contributes little to solving our counterfactual questions.
This highlights the need for novel prompting techniques for eliciting counterfactual reasoning ability or new paradigms for fine-tuning end-to-end multi-modal models to help with the counterfactual reasoning tasks.


\paragraph{Performance of Neuro-symbolic Models.}
Similar to end-to-end models, the performance of neuro-symbolic models on counterfactual questions is worse than that on the original ones. 
And it still holds that reasoning difficulty influences accuracy.
In the evaluation of ViperGPT~\cite{surismenon2023vipergpt}, the accuracy drops $12.4\%$ when one-step reasoning is required (\ie, `Numerical direct'), and it drops $48.4\%$ when multi-step reasoning is needed (\ie, `Numerical indirect'). 
We note that the gap of neuro-symbolic models between counterfactual performance and original performance is much larger than that of any end-to-end model on `Numerical indirect' and `Boolean' questions. 
This indicates that despite the fact that neuro-symbolic models like ViperGPT~\cite{surismenon2023vipergpt} and VisProg~\cite{Gupta2022VisProg} directly generate codes, it cannot handle complex counterfactual reasoning tasks.


\begin{mdframed}[backgroundcolor=blue!20] 
\noindent\textbf{Finding 1:} 
Performance of neuro-symbolic models are worse than end-to-end models on complex reasoning tasks.
\end{mdframed}

\begin{table}[t]
\caption{Evaluation on the GPT-4V model using a subset of \Real. Note that the results are not directly comparable with other tables.} \label{tab:gpt4v}
\vspace{-.8em}
\centering
\setlength{\tabcolsep}{11.5pt}
\begin{tabular}{lcc}
    \toprule
    GPT-4V               &  Original  &  Counterfactual\\
    \midrule
    Numerical Direct     &  50.4      &    14.6$\,_\text{\bf\textcolor{BrickRed}{(-35.8)}}$          \\
    Numerical Indirect   &  56.8      &    40.7$\,_\text{\bf\textcolor{BrickRed}{(-16.1)}}$        \\
    Boolean              &  65.9      &    48.0$\,_\text{\bf\textcolor{BrickRed}{(-17.9)}}$        \\
    \bottomrule
\end{tabular}
\end{table}

\begin{table*}[t]
\caption{Evaluation results of ViperGPT on \Real~when combining with different code LLMs.}
\label{tab:symbolic}
\vspace{-.8em}
\centering
\setlength{\tabcolsep}{8pt}
\begin{tabular}{lcccccc}
\toprule
\multirow{2}{*}{LLM of ViperGPT} & \multicolumn{2}{c}{Numerical Direct} & \multicolumn{2}{c}{Numerical Indirect} & \multicolumn{2}{c}{Boolean} \\

\cmidrule(r){2-3} \cmidrule(r){4-5} \cmidrule(r){6-7}
& Original & Counterfactual & Original  & Counterfactual & Original  & Counterfactual  \\

\midrule
GPT-3.5-turbo  &  \textbf{83.8} &  71.4$\,_\text{\bf\textcolor{BrickRed}{(-12.4)}}$ & \textbf{78.6}
 & \textbf{30.2}$\,_\text{\bf\textcolor{BrickRed}{(-48.4)}}$ & \textbf{95.0}  & 28.2$\,_\text{\bf\textcolor{BrickRed}{(-66.8)}}$ \\

CodeLlama-7B-Instruct  & 78.0 &  50.8$\,_\text{\bf\textcolor{BrickRed}{(-27.2)}}$ & 72.2
 &  20.4$\,_\text{\bf\textcolor{BrickRed}{(-51.8)}}$ & 70.5  &  31.9$\,_\text{\bf\textcolor{BrickRed}{(-38.6)}}$ \\
 
CodeLlama-13B-Instruct  &  78.9 &  \textbf{72.4}$\,_\text{\bf\textcolor{BrickRed}{(-6.5)}}$ & 73.4
 & 26.2$\,_\text{\bf\textcolor{BrickRed}{(-47.2)}}$ & 74.5  & \textbf{41.6}$\,_\text{\bf\textcolor{BrickRed}{(-32.9)}}$ \\

 CodeLlama-34B-Instruct  & 78.3  & 68.3$\,_\text{\bf\textcolor{BrickRed}{(-10.0)}}$ & 73.5 & 20.9$\,_\text{\bf\textcolor{BrickRed}{(-52.6)}}$ &  65.0 & 36.5$\,_\text{\bf\textcolor{BrickRed}{(-28.5)}}$ \\
 
WizardCoder-Python-7B  & 72.8 & 59.1$\,_\text{\bf\textcolor{BrickRed}{(-13.7)}}$ & 
68.6 & 18.3$\,_\text{\bf\textcolor{BrickRed}{(-50.3)}}$ &   27.7 & 29.5$\,_\text{\bf\textcolor{darkgreen}{(+1.8)}}$ \\

WizardCoder-Python-13B  & 73.7 & 62.3$\,_\text{\bf\textcolor{BrickRed}{(-11.4)}}$ & 
67.6 & 18.8$\,_\text{\bf\textcolor{BrickRed}{(-48.8)}}$ &  56.3  & 34.4$\,_\text{\bf\textcolor{BrickRed}{(-21.9)}}$ \\

WizardCoder-15B  & 78.0 &  55.4$\,_\text{\bf\textcolor{BrickRed}{(-22.6)}}$ & 
72.6 & 18.8$\,_\text{\bf\textcolor{BrickRed}{(-53.8)}}$ & 57.7  & 15.6$\,_\text{\bf\textcolor{BrickRed}{(-42.1)}}$ \\

\bottomrule
\end{tabular}
\vspace{-1em}
\end{table*}

\paragraph{Results of Neuro-symbolic Models With Different Code LLMs.}
Keeping the API tools, in-context prompts, and questions the same, we can test the coding abilities of different code-generation LLMs by replacing the ChatGPT model inside ViperGPT.
The evaluation results are shown in~\cref{tab:symbolic}.
Notably, the counterfactual presupposition causes significant and consistent performance drops across different model families and scales.
This observation emphasizes the inherent limitations of current language models in handling more complex reasoning scenarios.
Within this general trend, individual model performance varies. 
GPT-3.5-turbo generally outperforms other models in the original setting but also suffers from substantial drops in the counterfactual scenarios. 
CodeLlama~\cite{roziere2023code} shows a relatively moderate performance decrease when counterfactuals are introduced, suggesting some resilience but still a noticeable drop. 
WizardCoder~\cite{luo2023wizardcoder} performs least effectively in both original and counterfactual contexts.
Our results here highlight an important and urgent need for improvement in enhancing the models' abilities to handle counterfactual reasoning.

From the above experiments, we can see that no model families, neither end-to-end nor neuro-symbolic, can handle counterfactual questions well.
This indicates that \Dataset~is challenging and calls for further explorations.

\begin{mdframed}[backgroundcolor=blue!20] 
\noindent\textbf{Finding 2:} No model family can consistently address counterfactual questions.
\end{mdframed}

\paragraph{Evaluation of GPT-4V.}
We also provide an additional evaluation of our proposed benchmark with the GPT-4V~\cite{openai2023gpt4v}.
Due to the rate limit and the late date when the API becomes available, we only test GPT-4V on a $10\%$ randomly selected subset of \Real, results are presented in~\cref{tab:gpt4v}.
We show that despite the strong performance of GPT-4V at various visual problems~\cite{yang2023dawn}, our counterfactual questions still pose a challenge to GPT-4V as the model shows an over $10\%$ drop in performance for all three types of counterfactual questions in our benchmark.

\begin{mdframed}[backgroundcolor=blue!20] 
\noindent\textbf{Finding 3:} Even strongest models such as GPT-4V cannot solve our benchmark. 
\end{mdframed}

\begin{table}[t]
\caption{Evaluation results on~\Syn.} \label{tab:syn}
\vspace{-.8em}
\centering
\begin{tabular}{lcc}
\toprule
Model                             & Original & Counterfactual \\
\midrule
InstructBLIP (Vicuna-7B)          & 26.9      & 25.7            \\
InstructBLIP (FlanT5$_{\rm XXL}$) & 16.7     & 19.9           \\
BLIP2 (FlanT5$_{\rm XXL}$)        & 17.5      & 20.2            \\
\bottomrule
\end{tabular}
\end{table}

\subsection{Experiments on \bf{\Syn}}
We test the synthetic questions in the same way as the manually annotated questions. 
However, we find that almost no models show normal performance in the synthetic dataset \Syn.
As shown in~\cref{tab:syn} , the performance of InstructBLIP (Vicuna-7B) is at a random level $25\%$ and the performance of InstructBLIP (FlanT5$_{\rm XXL}$) and BLIP2 (FlanT5$_{\rm XXL}$) is worse than the random level.
We further prompt InstructBLIP (Vicuna) to merely count the flowers and it gives random answers, which indicates that it cannot recognize flowers in the synthetic images.
Similar to the observation of~\cite{zong2023fool}, we notice that the distribution of the options answered by these models is highly biased. 
For example, BLIP2 (FlanT5$_{\rm XXL}$) answers $(D)$ to most questions. 
It is important to emphasize that the prompt for multiple-choice questions must be meticulously designed; failure to do so may result in some models being unable to select any of the provided options.
For neuro-symbolic models, they perform better in the reasoning part but fail to answer the synthetic questions because no available APIs can handle the queries to the synthetic data. 
For example, ViperGPT cannot check whether a flower is in a polygon because no tools are available for querying this.
However, it is worth noting that it can often produce correct code even when counterfactual presuppositions are added. 
These results indicate that current models cannot generalize beyond the training domain of real images, let alone handling counterfactual reasoning in the OOD synthetic domain.

\begin{figure}[t]
    \centering
    \includegraphics[width=1\linewidth]{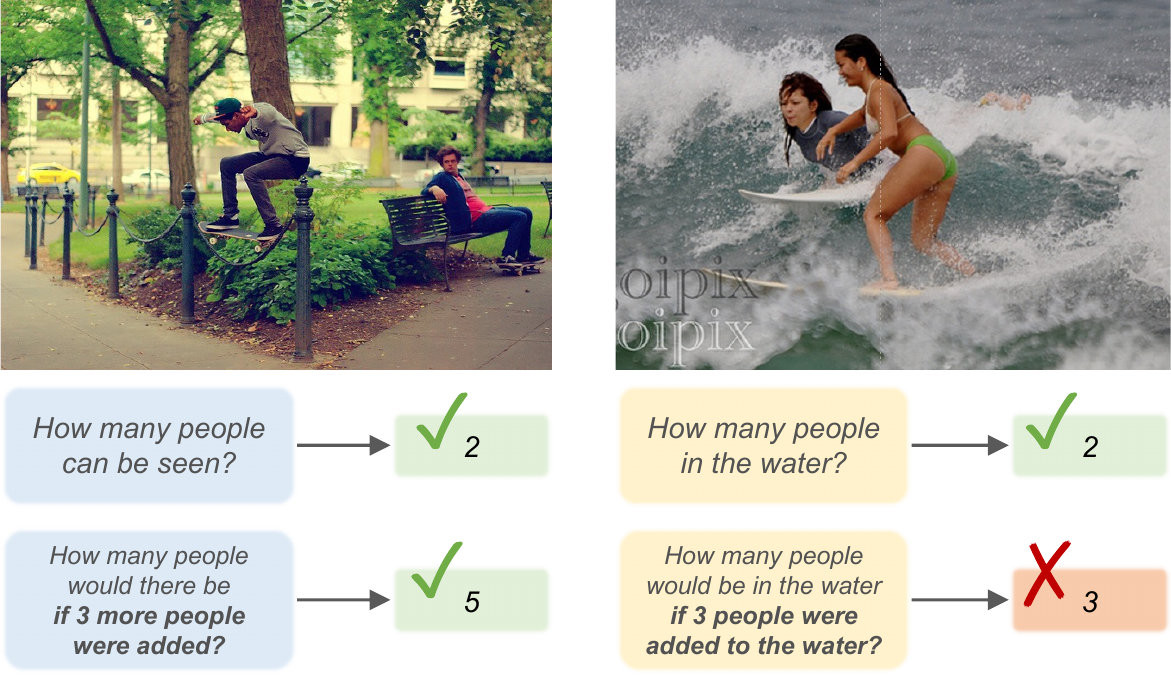}
    \vspace{-1em}
    \caption{\textbf{Qualitative example of biases in MLLMs}. Given similar questions, InstructBLIP (Vicuna-7B) provides correct answers for the male instance but incorrect answers for the female instance.}
    \label{fig:gender}
    \vspace{-1.5em}
\end{figure}

\subsection{Bias Analysis}
\begin{figure}[t]
    \centering
    \includegraphics[width=1\linewidth]{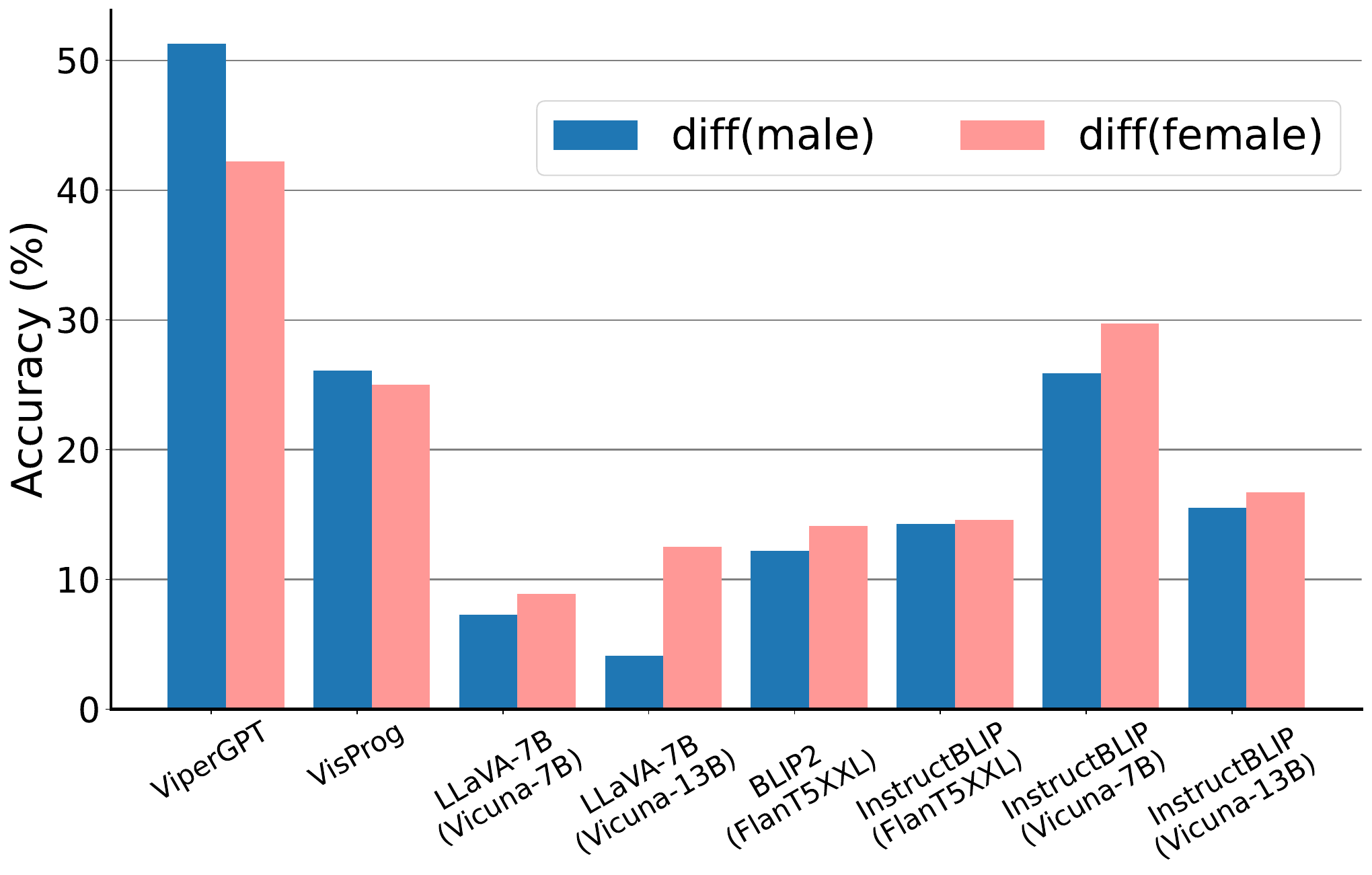}
    \vspace{-2em}
    \caption{\textbf{Performance difference of original and counterfactual questions on the male and female subgroup on \Real}. We can see that end-to-end models are often biased toward the male subgroup, and neuro-symbolic models are biased toward the female subgroup. The larger the gap between the performance differences, the larger the bias.}
    \label{fig:bias}
    \vspace{-1.5em}
\end{figure}

As our dataset \Real~is based on VQAv2 and the images were from the COCO dataset, we further study the bias of the COCO dataset in the counterfactual questions. 
Prior work~\cite{zhao2021captionbias} studies bias propagation pathways within image captioning on the COCO dataset by annotating the perceived gender and skin color of $28,315$ of the depicted people.
In virtue of this dataset, we could get the overlap between our dataset and the selected images and find that $985$ images in \Real~are annotated with gender and skin information. 
Among these images, we only keep those that contain only males or females, resulting in $717$ images eventually ($510$ male images and $207$ female images).  We then evaluate the performance of different models by subgroups. 
The result is shown in~\cref{fig:bias}.

Specifically, for each model we obtain the performance of both original and counterfactual questions of different subgroups.  
We then compute the difference between the accuracy of original and counterfactual questions in each subgroup.
Formally, we have the formula $ \texttt{diff(male)} = \texttt{male(ori)} - \texttt{male(cf)} $ and $ \texttt{diff(female)} = \texttt{female(ori)} - \texttt{female(cf)} $.
As shown in~\cref{fig:bias}, most end-to-end models have a larger $\texttt{diff(female)}$ than $\texttt{diff(male)}$, which indicates that when presenting counterfactual presuppositions, they struggle more with females images. Further studies are needed to understand the reasons and improve the fairness in the model reasoning process.

\begin{mdframed}[backgroundcolor=blue!20] 
\noindent\textbf{Finding 4:} MLLMs demonstrate systematic bias in answering gender-related counterfactual questions.
\end{mdframed}



\vspace{-1em}
\section{Conclusion}

In this paper, we study the ability of current multi-modal language models to handle counterfactuals -- a core cognition ability of human intelligence.
We build a novel dataset \Dataset~to test this counterfactual reasoning ability, the dataset is designed to have three types of counterfactual questions, with mixed real and synthetic images and questions.
Our evaluations on \Dataset~reveal several findings, the most significant finding is that no model in the current multi-modal model literature can consistently handle our counterfactual questions.
We have released our code and the dataset to help the community move forward on achieving human-level multi-modal intelligence.

{
\small
\bibliographystyle{ieeenat_fullname}
\bibliography{main}
}
\clearpage
\appendix
\onecolumn
\section{Overview}

In this supplementary material, we present more details and results.

\begin{itemize}
    \item We provide the prompts for ChatGPT to generate questions in \textbf{numerical direct} group and \textbf{boolean} group. 
    \item We show some qualitative results of end-to-end models. 
    \item We show some qualitative results of ViperGPT on~\Syn.
    \item  We present more detailed statistics and the top words for nouns and verbs of~\Dataset.
\end{itemize}

\section{Prompt for ChatGPT}
    In the process of annotating~\Real, we prompt ChatGPT to generate most new counterfactual modified questions for \textbf{numerical direct group} and \textbf{boolean group}. 
    to maximize the correctness of ChatGPT-generated questions,  we leverage chain-of-thought~\cite{wei2023chainofthought} strategy and insert in-context-examples~\cite{dong2023survey} into the prompt. We adopt different prompt patterns for the two groups, and the whole prompts are shown below.
\subsection{Numerical Direct Group}
    The counterfactuals for questions in numerical direct group are simple, so we prompt ChatGPT to produce the counterfactual suppositions straight. 
    Then the new answer can be obtained through simple calculations.

\begin{lstlisting}
You will change some numerical questions.
Your task is to perform the following actions:
1 - Read the original numerical question and answer
2 - Increase or decrease the number of items directly.
3 - Work out how this would change the answer to the question.
4 - Write a new question that asks how many items would there be if the number of items was increased or decreased according to the step 2.
    Change the original questions to new questions of unreal conditions with counterfactual presuppositions, using if clauses. Do not change the meaning of questions in new questions.
5 - Write the new answer to the new question.

Answer each initial question with the following format:
Original question:<original question>
Original answer:<original answer>
Step1:Add or remove <number> <item> to the original question.
Step2:<how the answer would change>
New question:<new question>
New answer:<new answer with a single number>

Here are some examples:
-----
Original question:How many birds are there?
Original answer:3
Step1:Add 3 birds
Step2:The answer would be 3+3=6
New question:How many birds would there be if 3 birds came?
New answer:6
-----
Original question:How many people in the picture?
Original answer:2
Step1:add 2 women
Step2:The answer would be 2+2=4
New question:How many people would be in the picture if there were 2 more women?
New answer:4
-----
Original question:How many zebras are here?
Original answer:2
Step1:1 zebra left and 2 zebras came
Step2:The answer would be 2-1+2=3
New question:How many zebras would there be if 1 zebra left and 2 zebras came?
New answer:1
-----
Original question:How many bikes are outside?
Original answer:2
Step1:double the bikes
Step2:The answer would be 2*2=4
New question:How many bikes would there be if the number of bikes doubled?
New answer:4
-----
Original question:How many sinks?
Original answer:2
Step1:add 2 sinks
Step2:The answer would be 2+2=4
New question:How many sinks if two more sinks were added?
New answer:4
-----
Original question:How many oranges are there?
Original answer:2
Step1:eat all oranges
Step2:The answer would be 2-2=0
New question:How many oranges would there be if all oranges were eaten?
New answer:0
-----
Original question:How many animals are here?
Original answer:2
Step1:another zebra comes
Step2:The answer would be 2+1=3
New question:How many animals would there be if another zebra came?
New answer:3
-----
Original question:How many birds are there?
Original answer:3
Step1:2 birds fly away
Step2:The answer would be 3-2=1
New question:How many birds would there be if 2 birds flew away?
New answer:1
-----

Now change the following questions step by step:
\end{lstlisting}

\subsection{Boolean Group}
To ensure the model fully understands the counterfactual suppositions, we propose that the new answers should be different from the original ones.
However, applied with the same prompt strategy as numerical direct group, ChatGPT often fails to flip the original answer for questions in boolean group. 
Therefore we alter the CoT strategy as follows:
\begin{itemize}
    \item Flip the original answer and describe what the situation is now.
    \item Design a counterfactual supposition that can make this situation true.
\end{itemize}
\begin{lstlisting}
You will change some questions.
Your task is to perform the following actions:
1 - Read the original yes/no question and answer
2 - FLip the original Answer
3 - Work out how to make the answer true.
4 - Write a new question that asks if the answer would be true if the action you worked out in step 3 was performed.
    Change the original questions to new questions of unreal conditions with counterfactual presuppositions, using if clauses. Do not change the meaning of questions in new questions.
5 - Write the new answer to the new question.

Answer each initial question with the following format:
Original question:<original question>
Original answer:<original answer>
Step1:The new answer should be <yes/no>, so ...
Step2:How to make ...:...
New question:<new question>
New answer:<new answer>

Here are some examples:
-----
Original question:Are the goggles covering her eyes?
Original answer:yes
Step1:The new answer should be no, so the goggles are not covering her eyes.
Step2:How to make the goggles not cover her eyes: take off the glasses.
New question:Would the goggles be covering her eyes if she took off the glasses?
New answer:no
-----
Original question:Is there a hotdog on this car?
Original answer:yes
Step1:The new answer should be no, so there is no hotdog on this car.
Step2:How to make there be no hotdog on this car: remove all food.
New question:Would there be a hotdog on this car if all food was removed?
New answer:no
-----
Original question:Are these vegetables cooked?
Original answer:yes
Step1:The new answer should be no, so these vegetables are not cooked.
Step2:How to make these vegetables not be cooked: make them raw.
New question:Would these vegetables be cooked if they were raw?
New answer:no
-----
Original question:Is he happy?
Original answer:no
Step1:The new answer should be yes, so he is happy.
Step2:How to make him happy: make him laugh.
New question:Would he be happy if he was laughing?
New answer:yes
-----
Original question:Is this woman doing something active?
Original answer:no
Step1:The new answer should be yes, so she is doing something active.
Step2:How to make her do something active: make her dance.
New question:Would this woman be doing something active if she was dancing?
New answer:yes
-----
Original question:Is the ground wet?
Original answer:no
Step1:The new answer should be yes, so the ground is wet.
Step2:How to make the ground wet: make it rain.
New question:Would the ground be wet if it was raining?
New answer:yes
-----
Original question:Is the sky clear?
Original answer:yes
Step1:The new answer should be no, so the sky is not clear.
Step2:How to make the sky not clear: make it cloudy.
New question:Would the sky be clear if it was cloudy?
New answer:no
-----
Original question:Is the plane flying?
Original answer:no
Step1:The new answer should be yes, so the plane is flying.
Step2:How to make the plane fly: make it take off.
New question:Would the plane be flying if it took off?
New answer:yes
-----

Now change the following questions step by step:
\end{lstlisting}

\section{Qualitative Result of End-to-end Models}
\begin{figure}[ht]
    \centering
    \includegraphics[width=.85\linewidth]{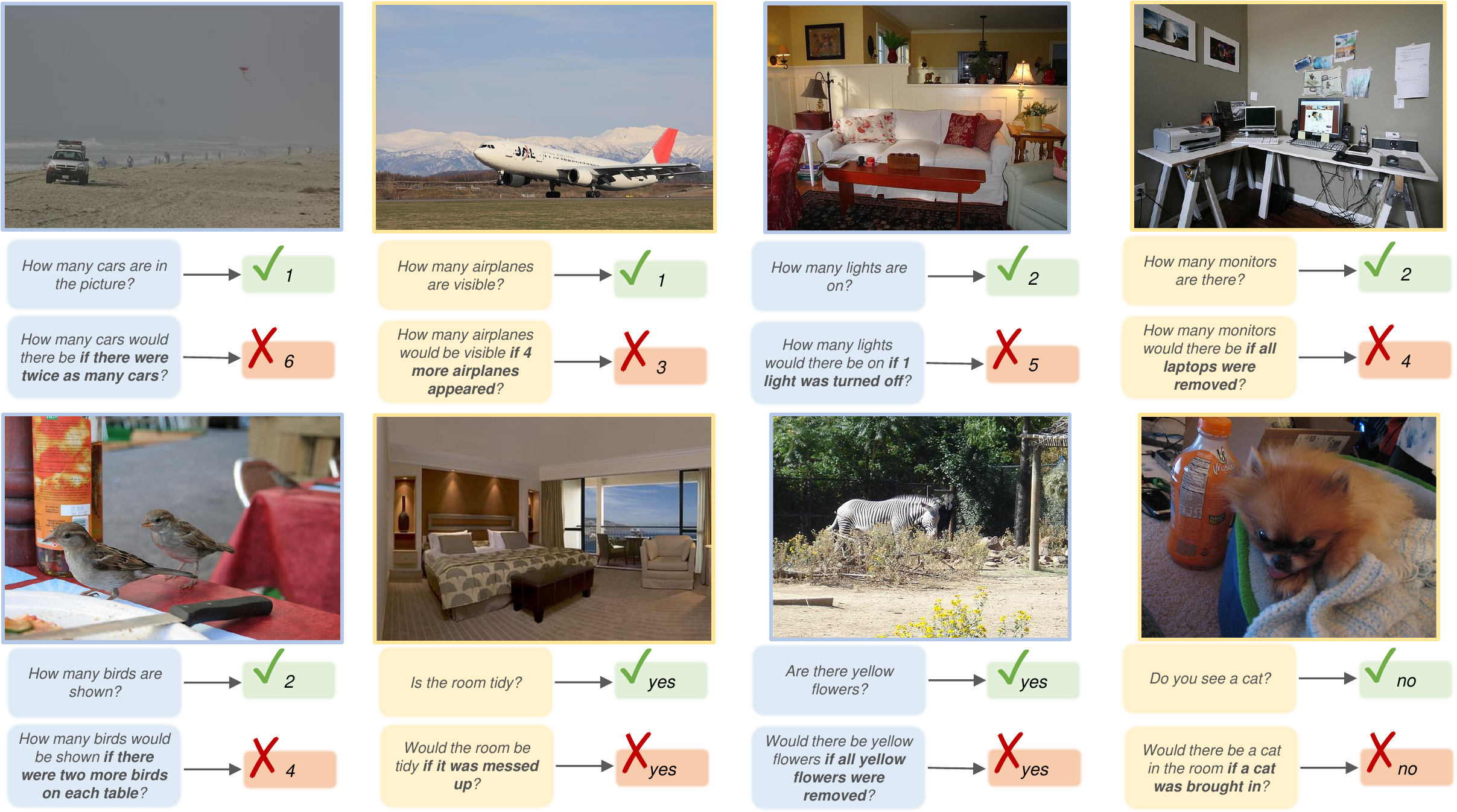}
    \caption{Common failure cases of end-to-end models on~\Dataset, the added counterfactual presupposition are in \textbf{bold}. }
    \label{fig:end1}
\end{figure}

\begin{figure}[h]
    \centering
    \includegraphics[width=.85\linewidth]{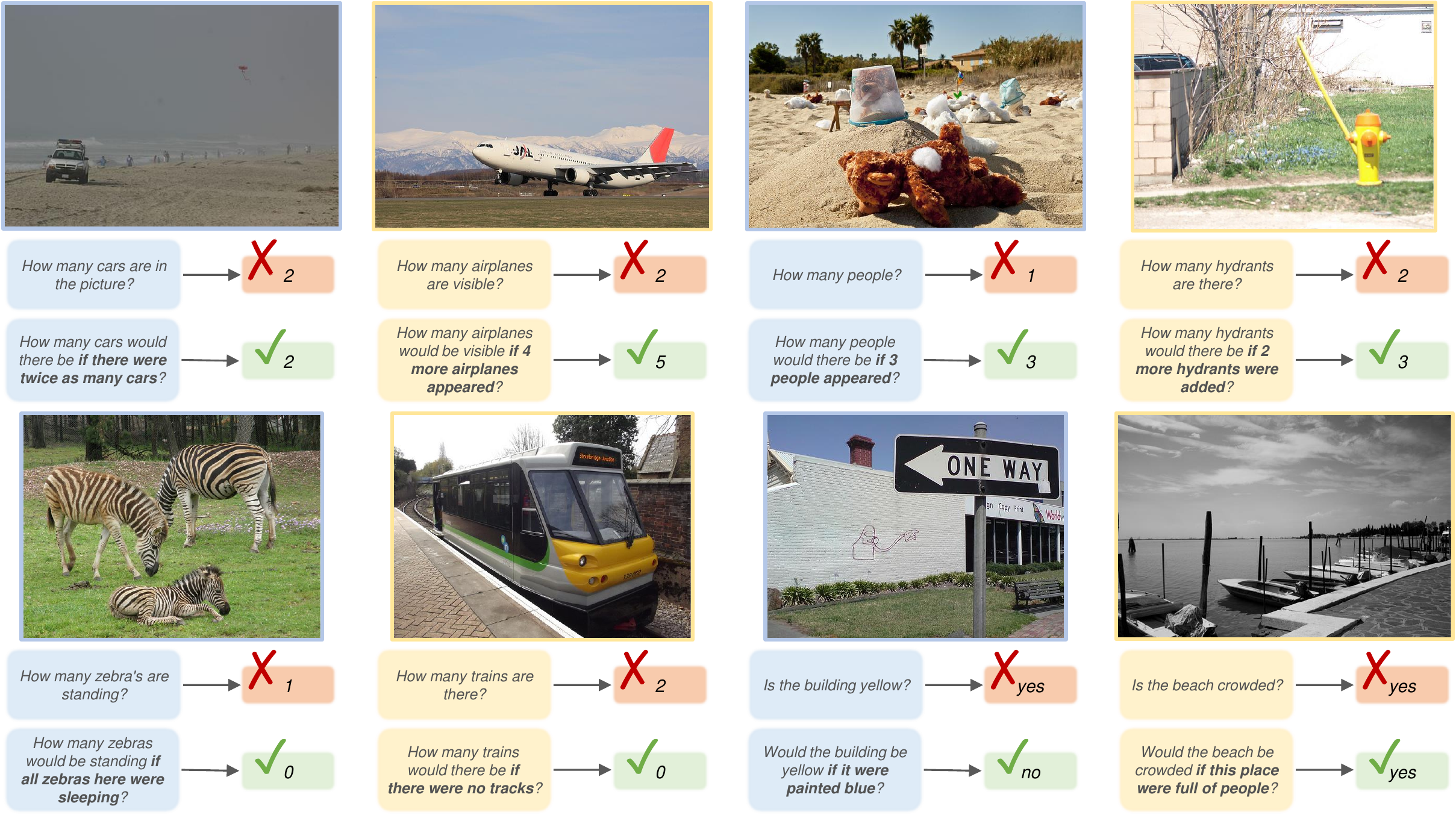}
    \caption{\textbf{Examples of LLaVA-13B.} The model answers the counterfactual questions correctly but fails at the original questions.}
    \label{fig:end2}
\end{figure}

When counterfactuals are added, most models fail to provide correct answers and examples are provided in~\cref{fig:end1}. 
We also notice that there exists some weird data in the result table. 
For example, LLaVA-13B (Vicuna-13B)~\cite{liu2023llava} gets 31.2\% correct for both original and counterfactual questions in numerical direct group. 
We inspect its result and find out that it is often the case that LLaVA-13B (Vicuna-13B) answers the counterfactual questions correctly but answers the original questions incorrectly.
Several instances are shown in~\cref{fig:end2}.

\begin{figure}[h]
\centering

\begin{minipage}{.45\linewidth}
\setlength{\tabcolsep}{.5mm}
\tablestyle{4.6pt}{1.2}
\begin{tabular}{cc}
\toprule
\textbf{Type}                                   & \textbf{Counts} \\ 
\midrule
Type-Token Ratio                       & 0.1251                     \\
Verb-Token Ratio (total \# verb-types) & 0.1479                     \\
Verb-Token Ratio (total \# types)      & 0.0287                     \\
Noun-Token Ratio (total \# noun-types) & 0.1587                     \\
Noun-Token Ratio (total \# types)      & 0.0783                     \\
Direct/Indirect/Boolean (\%)           & 37/27/36                   \\
\bottomrule
\end{tabular}

\end{minipage}
\begin{minipage}{.22\linewidth}
\includegraphics[width=\textwidth]{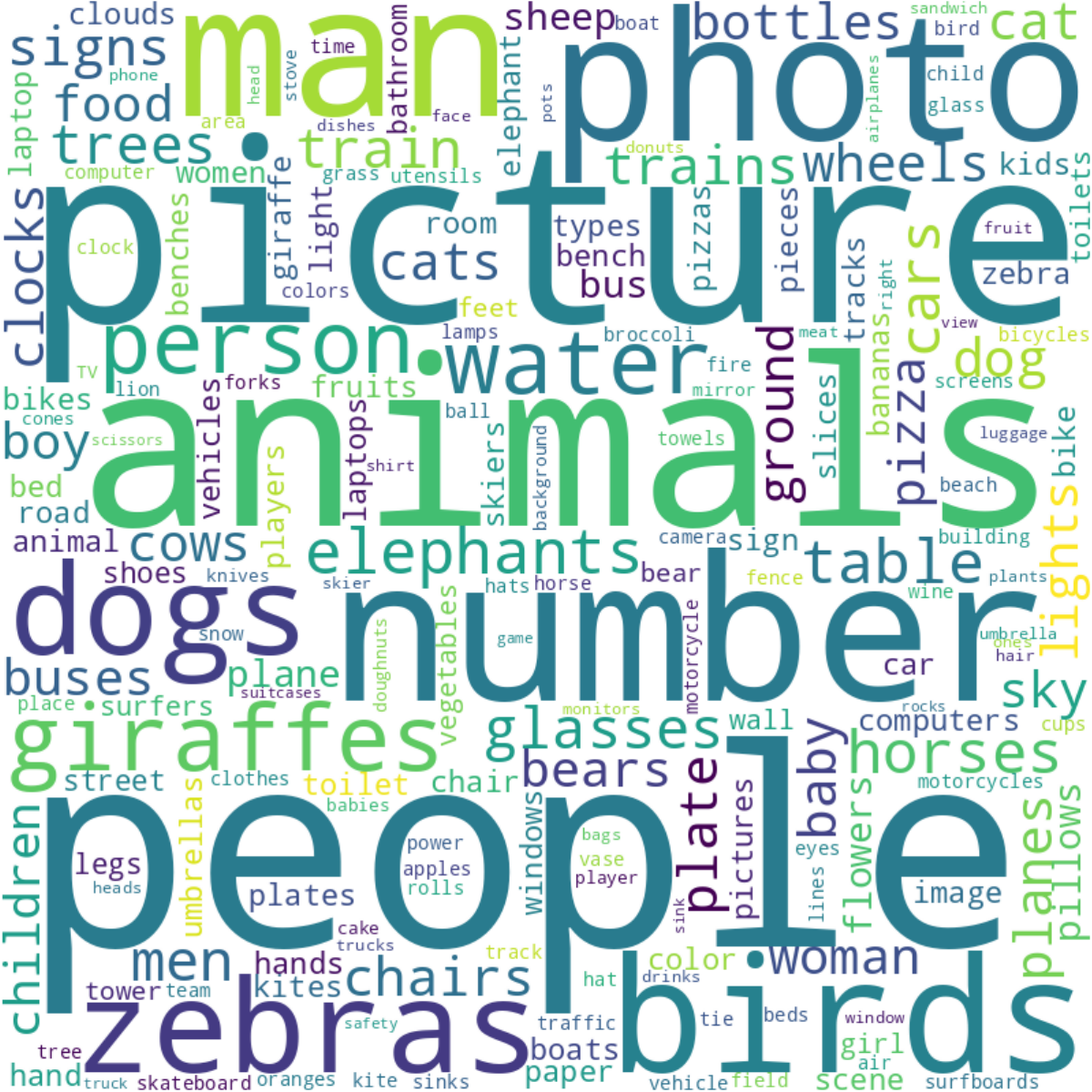}
\end{minipage}
\hspace{.8em}
\begin{minipage}{.22\linewidth}
\includegraphics[width=\textwidth]{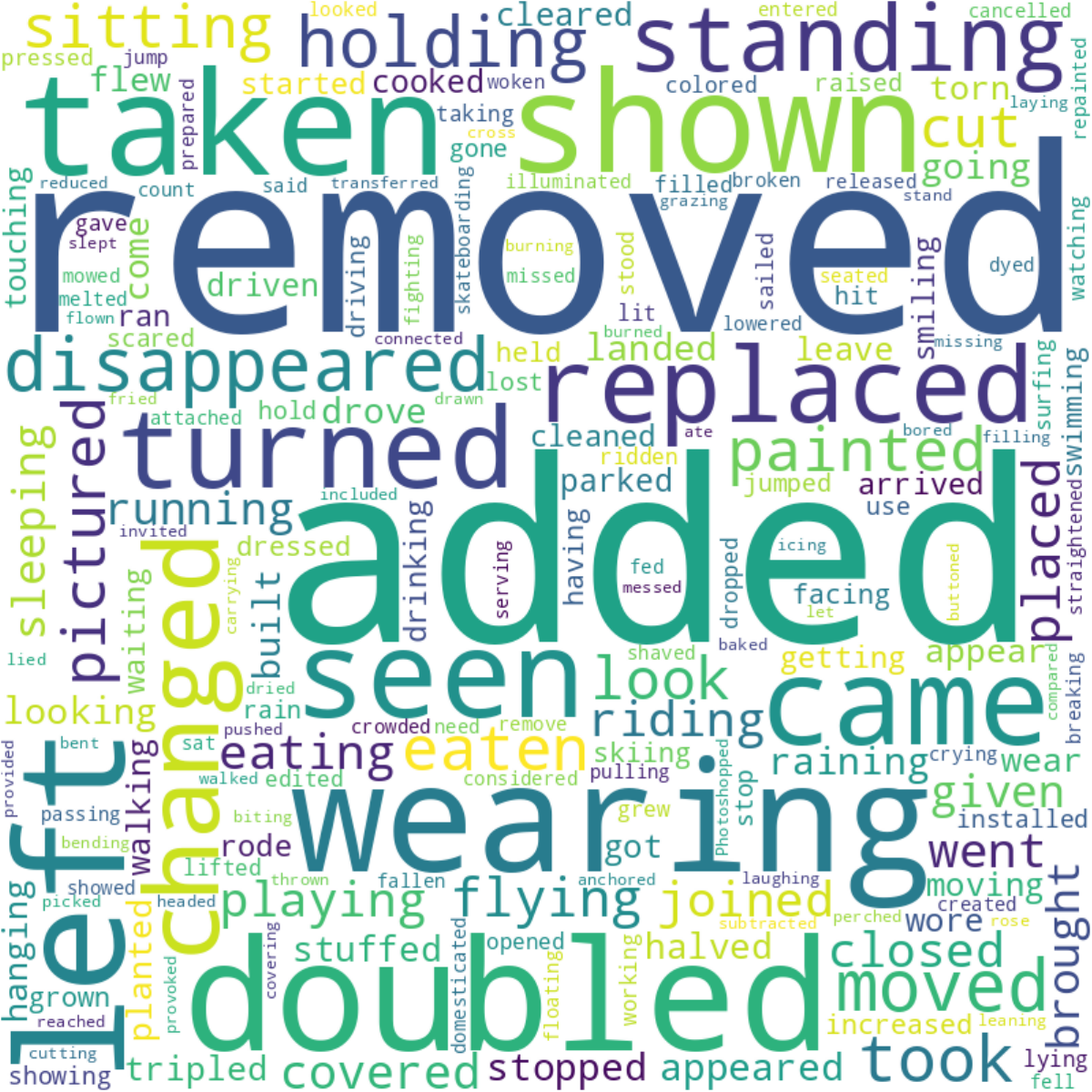}
\end{minipage}

\caption{\textbf{More detailed statistics of~\Dataset}. The table shows the type-token ratio and the distribution of counterfactual questions. The pictures are word clouds of nouns and verbs in~\Dataset.}
\label{fig:ttr}
\end{figure}

\section{Qualitative Result of Neuro-symbolic Models}
When evaluating ViperGPT~\cite{surismenon2023vipergpt}, We inspect the codes generated by ChatGPT.
The codes are sometimes wrong in that ChatGPT misunderstands or even entirely ignores the counterfactuals.

\begin{figure}[h]
    \centering
    \includegraphics[width=1\linewidth]{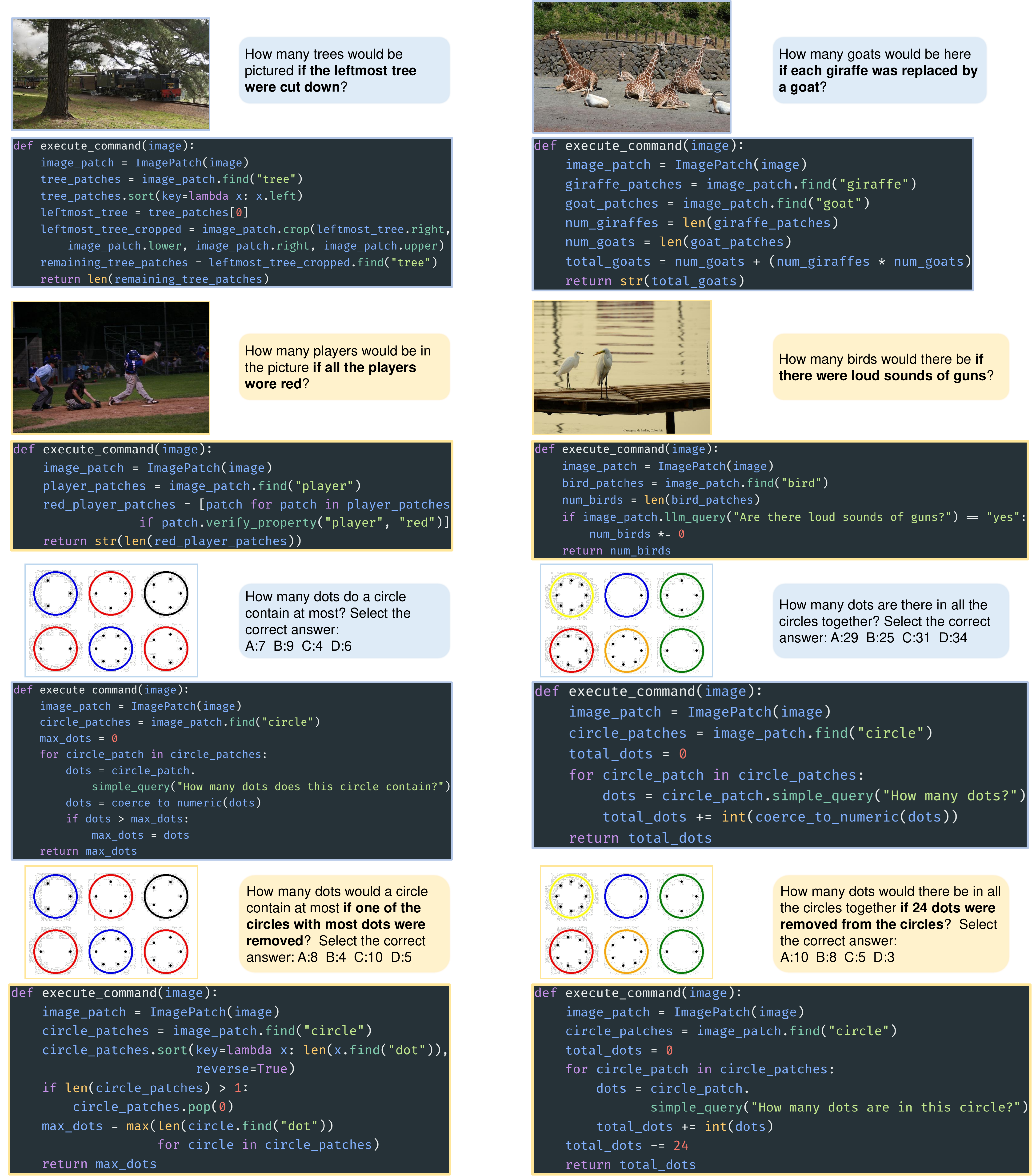}
    \caption{\textbf{The codes generated by ViperGPT}. Here the codes for~\Real~are logically wrong, and the codes for~\Syn~are logically correct.}
    \label{fig:vipergpt}
    \vspace{-1.5em}
\end{figure}

ViperGPT fails to handle~\Syn~in that it cannot check whether a flower is in a polygon. 
Despite this, the codes generated for the~\Syn~is inspiring.
We notice that it can often produce correct code even when counterfactual presuppositions are added. 
Some example codes are provided in~\cref{fig:vipergpt}.

\section{Additional Detailed Statistics}
The type-token ratios(TTR) of nouns and verbs are provided in~\cref{fig:ttr}. The TTR of~\Dataset~is higher than other common datasets in that the words in presuppositional statements are highly duplicated with that in main clauses. The word clouds in~\cref{fig:ttr}~reveal the top words for nouns and verbs in~\Dataset.

\section{Limitations, future works, and broader impact}

Although our work presents the first of its kind evaluation of counterfactual reasoning abilities of multi-modal large language models, the number of images in~\Dataset~is a limiting factor. 
However, with the 3K counterfactual questions on real images and the additional 3K images on synthetic data, we are able to show a drastic drop in performance when current state-of-the-art multi-modal models are evaluated on the counterfactual questions.
One future work we are planning is to gather annotations for more images and cover more models for the evaluation.

As counterfactual reasoning is considered as the corner stone of human intelligence, we hope our proposed benchmark could help evaluate the progress towards artificial general intelligence or building the next generation of AI assistant. 

\clearpage
\section*{Acknowledgement}
Yongshuo Zong was supported by the United Kingdom Research and Innovation (grant EP/S02431X/1), UKRI Centre for Doctoral Training in Biomedical AI at the University of Edinburgh, School of Informatics. For the purpose of open access, the author has applied a creative commons attribution (CC BY) licence to any author accepted manuscript version arising.

\end{document}